\documentclass[a4paper, 10pt, conference]{ieeeconf}
\usepackage{graphicx} % Required for inserting images
\usepackage{lipsum}
\usepackage{hyperref}
\usepackage{svg}
\IEEEoverridecommandlockouts
\title{Can a mobile robot learn from a pedestrian model to prevent the sidewalk salsa?}
\author{Olger Siebinga$^{1}$ and David Abbink$^{1}$
\thanks{$^{1}$Departments of Cognitive Robotics and Sustainable Design Engineering, Delft University of Technology, Delft, The Netherlands. Corresponding author: {\tt\small o.siebinga@tudelft.nl}}%
\thanks{\textbf{Data and Software Availability:}
The software and simulated data underlying this publication are available upon request}%
\thanks{This work was presented as a poster presentation at the workshop \textit{Movement Matters – A Turing Test for Robot Interaction} at the \textit{2025 IEEE International Conference on Robot \& Human Interactive Communication (RO-MAN)} in Eindhoven, the Netherlands}}
\date{June 2025}

\begin{document}

\maketitle

\begin{abstract}
    Pedestrians approaching each other on a sidewalk sometimes end up in an awkward interaction known as the "sidewalk salsa": they both (repeatedly) deviate to the same side to avoid a collision. This provides an interesting use case to study interactions between pedestrians and mobile robots because, in the vast majority of cases, this phenomenon is avoided through a negotiation based on implicit communication. Understanding how it goes wrong and how pedestrians end up in the sidewalk salsa will therefore provide insight into the implicit communication. This understanding can be used to design safe and acceptable robotic behaviour. In a previous attempt to gain this understanding, a model of pedestrian behaviour based on the Communication-Enabled Interaction (CEI) framework was developed that can replicate the sidewalk salsa. However, it is unclear how to leverage this model in robotic planning and decision-making since it violates the assumptions of game theory, a much-used framework in planning and decision-making. Here, we present a proof-of-concept for an approach where a Reinforcement Learning (RL) agent leverages the model to learn how to interact with pedestrians. The results show that a basic RL agent successfully learned to interact with the CEI model. Furthermore, a risk-averse RL agent that had access to the perceived risk of the CEI model learned how to effectively communicate its intention through its motion and thereby substantially lowered the perceived risk, and displayed effort by the modelled pedestrian. These results show this is a promising approach and encourage further exploration.
\end{abstract}

\section{Introduction}
\begin{figure*}
    \centering
    \includegraphics[width=.75\textwidth]{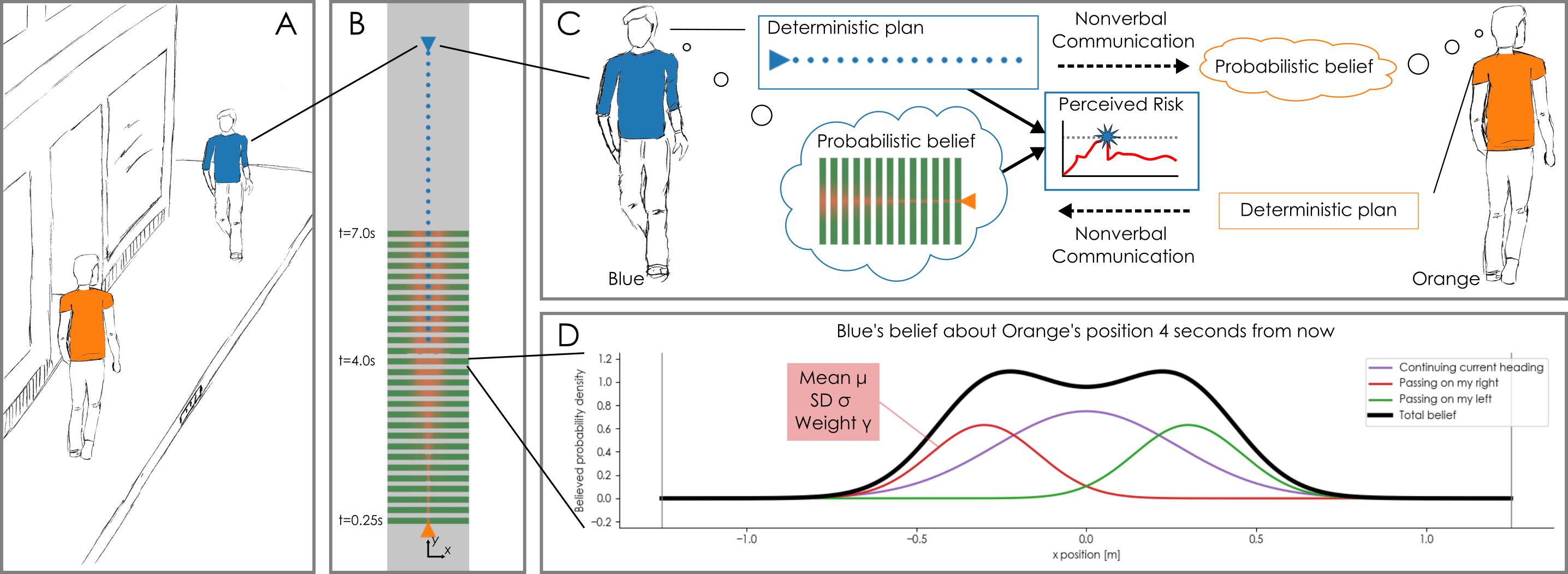}
    \caption{(reprinted from~\cite{Siebinga2025}) \textbf{Panel A:} Two pedestrians approaching each other on the sidewalk and potentially ending up in the sidewalk salsa. \textbf{Panel B:} a top-down view representing the key components of the Communication-Enabled Interaction model presented in~\cite{Siebinga2025}: the plan of the blue pedestrian (approach from the top) and its belief about the orange pedestrian. \textbf{Panel C:} The model structure of a CEI model consists of this deterministic plan and probabilistic belief for both pedestrians. Combining these results in a perceived risk, when this risk exceeds a personal threshold, pedestrians unilaterally change their plan to lower the risk. Pedestrians base their belief on implicit communication they observe, in this case (lateral) position and heading. \textbf{Panel D:} This belief consists of a probability distribution over lateral positions for a fixed point in time in the future. If the other pedestrian is walking on the left side of the sidewalk and/or heading towards that side, it is believed they are more likely to pass on that side. This way, position and heading cues are observed as implicit communication and result in a belief. }
    \label{fig:intro}
\end{figure*}

% - salsa intro
Suppose you're walking down a sidewalk and another pedestrian comes around a corner and approaches you head-on (Figure~\ref{fig:intro}-A). You move to one side to avoid them, but they simultaneously move to the same side. When you try to move to the other side of the sidewalk, they do the same. The third attempt to move works, and after an awkward interaction, you can both continue.

% - role of communication
Although you will probably not experience this every day, it is a well-known phenomenon known under many names~\cite{Siebinga2025}. Here, we refer to it as the "Sidewalk Salsa". This phenomenon provides an interesting use case to study interactions between pedestrians and mobile robots since it encompasses a negotiation based on implicit communication to come to a safe solution (i.e., not colliding). In the vast majority of human-human interactions, this negotiation leads to a successful and mutually accepted outcome. Yet, in the case of a sidewalk salsa, this negotiation fails; A miscommunication happens. If we understand how this failure occurs, we understand how we can prevent it. This knowledge can potentially help design communicative automated behaviour for robots operating in environments where they interact with individual pedestrians; for example, mobile parcel delivery robots (e.g.,~\cite{Gehrke2023, Weinberg2023}) or indoor mobile robots (e.g.,~\cite{Dadvar2021, Chen2018}) used in healthcare or industry~\cite{Fang_Mei_Yuan_Wang_Wang_Wang_2021, Möller_Furnari_Battiato_Härmä_Farinella_2021}.

% - CEI model
One way to gain a deeper understanding of human behaviour in these interactions is by making a computational model describing it~\cite{Guest2021}. One modelling framework specifically aims to describe human-human interactions where (implicit) communication plays an important role: the Communication-Enabled Interaction (CEI) framework~\cite{Siebinga2023}. In this framework, pedestrians (or other traffic participants) are assumed to have a deterministic plan for their own future positions and a probabilistic belief about others' future positions (Figure~\ref{fig:intro}-C). Combined, these result in a perceived risk (of colliding); when this risk exceeds a personal risk threshold, pedestrians are assumed to change their plan to lower the risk. Communication is what links one pedestrian's plan to another pedestrian's belief, i.e., pedestrians form a belief based on observed communication. Models based on this framework have successfully captured driving behaviour in merging interactions~\cite{Siebinga2024, siebinga_modeling_2025}, and reproduced the sidewalk salsa~\cite{Siebinga2025}.

% - gap of how to use this in a robot
However, although the process of computational modelling can result in fundamental knowledge about human behaviour~\cite{Guest2021}, and this knowledge has the potential to inform robotic decision-making and path planning, it is unclear how to directly leverage such a model in robotic planning and decision-making algorithms. In automotive research, human behaviour models (i.e., driver models) have been used to improve autonomous behaviour. For example, by using models in benchmark tests for verification and validation~\cite{Queiroz2024, Li2018}. This approach allows for an indirect effect on robotic behaviour: the control algorithms can be (manually) improved after validation. Robotic behaviour changes through a design iteration of the algorithm; the model has no direct effect on the planning and decision-making. 

Some approaches use models that directly inform autonomous planning; mostly in the context of game theory (e.g.~\cite{Sadigh2018, Schwarting2019}). Although useful from a control perspective, game theory makes strong assumptions about human behaviour that are questionable in the context of traffic interactions~\cite{Siebinga2023}. Most importantly, game theory assumes that there is no communication between "players" since it is not in their best interest to inform others about their plans when aiming to win a game. How to use models that do not assume human rationality and allow for communication directly in robotic planning and discussion making is thus an under-explored topic.

% - idea of RL and paper outline
One possibility to directly leverage a behaviour model for robotic planning and decision-making is to train a Reinforcement Learning (RL) agent to control the robot while interacting with the model during training. Provided that the model is sensitive to the same communicative cues human pedestrians use while interacting, the RL agent might learn how to express these cues to safely and acceptably interact with pedestrians. An added benefit of this approach is that many models try to capture the internal representation humans use in their decision-making, i.e., models aim to capture the underlying mechanisms that drive human behaviour. When an RL agent has access to this internal state during training, it can potentially learn to influence how it is perceived in a positive way. For example, in the case of a CEI model, the RL agent could be trained to minimise the risk perceived by pedestrians it interacts with.

Here, we present an initial exploration of this approach, showing a proof-of-concept for training an RL agent in interaction with the CEI model of pedestrians~\cite{Siebinga2025}. We compare the RL agent with a social forces model~\cite{helbing_social_1995} and show that the agent successfully learns to express useful communicative cues. Furthermore, we train a second RL agent to minimise the perceived risk. This risk-averse agent learns a slightly different behaviour compared to the basic RL agent, yet its actions are more communicative towards the CEI model. With this proof-of-concept, we show a promising first step towards a new approach for leveraging human behaviour models in planning and decision-making for mobile robots.

\section{Methods}
We used the simulation environment\footnote{https://github.com/tud-hri/sidewalk-simulation} developed for the CEI model of the sidewalk salsa as a baseline for the reinforcement learning environment. The CEI model as described in~\cite{Siebinga2025} controlled one of the pedestrians in the simulation. These pedestrians are modelled with a dynamic pedestrian model~\cite{Mombaur2008}, which allows steering inputs (as in a bicycle model), yet also allows sidestepping in the form of lateral velocity and acceleration. The sidewalk in the simulation was $15~m$ long and $2.5~m$ wide. The modelled pedestrians interact with a simulated robot. We implemented the original social forces model as described in~\cite{helbing_social_1995} as a baseline robot. We used two-dimensional point mass dynamics with the same parameters as in the original model, but we excluded the noise, visual view angle, and attraction forces.

Furthermore, we trained two Reinforcement Learning agents in interaction with the CEI agent to control the robot. Both RL agents used the same architecture, training method, and reward function, except that for the risk-averse agent, we included a minor penalty in the reward function based on the perceived risk of the CEI agent. The policy network was a multilayer perceptron with three hidden layers of 256 neurons each, using LayerNorm and LeakyReLU activations. The network outputs the mean and standard deviation for normalised input for a dynamic bicycle model: acceleration and steering ([-1, 1]). During training, actions were sampled; during evaluation, we used the mean. The agents observed the full state of the robot and pedestrian (positions, velocities, headings, angular velocities) and the previous action taken by the robot.

Training was performed using the REINFORCE algorithm (AdamW optimiser, learning rate $1e-3$, discount factor $0.99$), in batches of 20 episodes, for a total of 12,000 episodes. We used curriculum learning, where in the first 1000 episodes, there was no pedestrian present, in the second 1000, the pedestrian was stationary, and after that, the pedestrian behaved according to the CEI model. The reward function provided small positive rewards for progress along the sidewalk and small penalties for deviation from the desired velocity and heading. Furthermore, small penalties were given for cumulative inputs and the input derivative. Collisions and going beyond track bounds were heavily penalised, and a large reward was given when the robot reached the opposite end of the sidewalk. 

During training, the CEI agent was assigned a random risk threshold, sampled from a uniform distribution $[0.6, 0.9]$. Both agents were placed at the centre of the sidewalk on opposite ends with a random offset sampled from a uniform distribution $[-0.4~m, 0.4~m]$ in $x$ and $y$ direction.

To evaluate the robotic agents, we simulated $200$ interactions with a simulated pedestrian for each robot (social forces, basic RL agent, and risk-averse RL agent). In this validation, the CEI model was assigned a random risk threshold sampled from a uniform distribution $[0.6, 0.7]$, consistent with the values used in~\cite{Siebinga2025}. Initial positions were appended with a random offset from a uniform distribution $[-0.1~m, 0.1~m]$, to increase the conflict with respect to training.

\section{Results}
\begin{figure}
    \centering
    \includesvg[width=.9\linewidth]{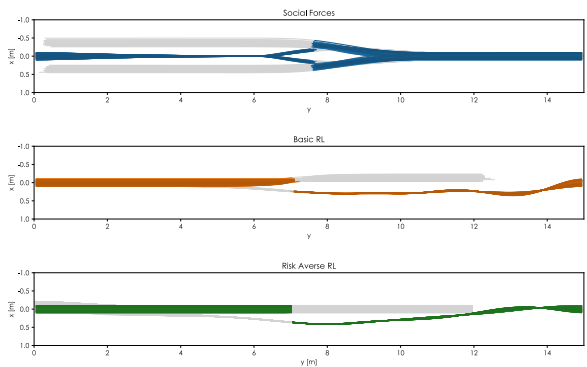}
    \caption{The position traces of the agents during evaluations. In all cases, the modelled pedestrian (CEI agent) approaches from the left side while to robot approaches from the right side. Coloured lines show the position traces up until the point where the agents pass each other. Grey lines show the remainder of the trajectories.}
    \label{fig:traces}
\end{figure}

All $600$ simulations ended with both the simulated pedestrian and robot reaching the opposite side of the sidewalk, avoiding collisions and going out of bounds. Both RL agents thus successfully learned to interact with the CEI model. The position traces (Figure~\ref{fig:traces}) show substantial differences between the simulated robots. While the social forces model follows the centre of the sidewalk until it enters the influence of the other pedestrians, the RL agents learned to pick a side early on. This clearly signals to the pedestrian on which side it will pass, thus the RL agent learned how to communicate its intention through interactions with the model. The risk-averse agent shows a subtle difference in behaviour that has a substantial effect on the simulated pedestrian.

The basic RL agent learned to pick a side to pass the pedestrian as quickly as possible. While this is already a good strategy to avoid collisions, the risk-averse agent learned that the CEI model uses heading angle as an important cue in communication~\cite{Siebinga2025}. Therefore, the risk-averse agent maintains its position at the sidewalk centre initially, until it reaches the range of influence of the pedestrian. The CEI model uses a mechanism that increases uncertainty for belief points in the future; thus, the influence of communication only increases with a certain look-ahead time~\cite{Siebinga2025}. Within this range (roughly between the $7$ and $12~m$ markers in Figure~\ref{fig:traces}), the robot steers clear of the pedestrian with a heading pointing away from the pedestrian (instead of parallel to the sidewalk in the basic RL agents case). This successfully communicates to the pedestrian that it is planning to pass on that side.

\begin{figure}
    \centering
    \includesvg[width=.8\linewidth]{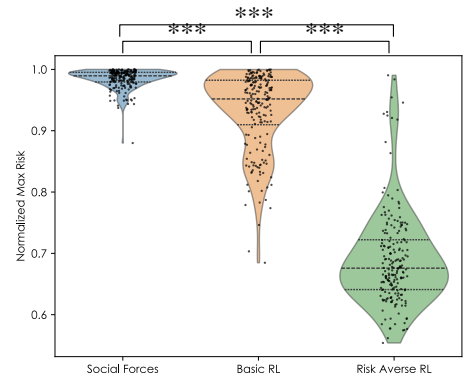}
    \caption{The normalised maximum risk perceived by the CEI agent while interacting with the different robots. A value of $1.0$ corresponds to the (randomised) risk threshold of the CEI agent. Dots show all datapoints, dashed horizontal lines indicate the interquartile ranges. A Kruskal–Wallis test revealed a significant difference between groups, $H(2)=427.55$, $p<.001$. Post-hoc tests with Bonferroni correction showed significant differences ($p<.001$) between all conditions.}
    \label{fig:risk}
\end{figure}

This successful communication is reflected in the risk perceived by the simulated pedestrian (Figure~\ref{fig:risk}). While almost all interactions with the social forces model result in the CEI agent's perceived risk reaching its threshold, the basic RL agent substantially lowers the perceived risk. The risk-averse agent shows an even lower maximum perceived risk; in this case, almost none of the CEI agents reach their risk threshold. The risk-averse agent successfully learned how to change its communicative behaviour to lower the perceived risk of a pedestrian.

\begin{figure}
    \centering
    \includesvg[width=.8\linewidth]{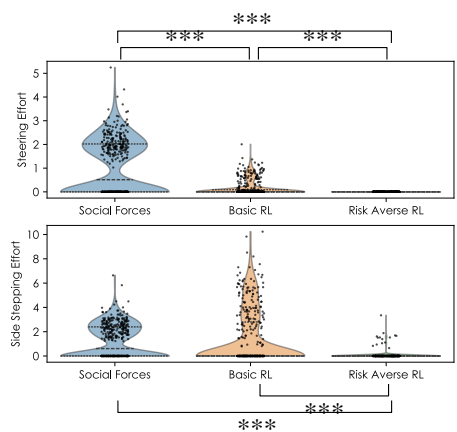}
    \caption{The cumulative inputs applied by the CEI agent while interacting with the different robots. Separated into steering inputs and sidestepping inputs. Dots show all datapoints, dashed horizontal lines indicate the interquartile ranges. Kruskal–Wallis tests revealed significant difference between groups for side stepping input ($H(2)=43.84$, $p<.001$) and steering ($H(2)=66.69$, $p<.001$). Post-hoc tests with Bonferroni correction showed significant differences ($p<.001$) between all conditions except Basic RL/Social Forces for side-stepping input ($p=1.0$).}
    \label{fig:effort}
\end{figure}

These lower risks in turn lead to lower effort, or input actions, taken by the pedestrian to mitigate a collision (Figure~\ref{fig:effort}). In interaction with the social forces model, the CEI model shows the highest median cumulative steering and side-stepping inputs. The basic RL agent causes the CEI model to use higher side-stepping actions in some simulations, yet both median inputs are lower compared to the social forces model. In the case of the risk-averse agent, the CEI model hardly uses any inputs at all. While the position traces (Figure~\ref{fig:traces}) of the two RL agents show only minor differences, this minor behavioural difference has a large effect on how the behaviour of the robot is understood by the modelled pedestrian and thus on how much action they take to lower their perceived risk.

\section{Discussion}
Although promising as a first step, the results of this work need to be interpreted with care. Our work illustrates the potential of our proposed approach, but without the rigorous work needed to conclude whether using a CEI model to train an RL robot is beneficial in real-world settings. The first major limitation in this work is that the CEI model used is not (yet) validated on pedestrian behaviour. So, although we can conclude that the RL agents have learned to successfully communicate their intention through their movement to the CEI model, this does not necessarily generalise to interaction with human pedestrians. 

A second major limitation is that we have used a heavily simplified simulation of interactions on a sidewalk. Both agents are represented by simple dynamical models and are assumed to have full control without (output) noise. The robot has access to full observations without noise. And the RL agents are only trained on a sidewalk with fixed dimensions, interacting with a single other pedestrian at the same velocity in every trial. Yet, even under these simplified conditions, the robot shows behaviour that might not be desirable in a real robot. Both RL agents show some swerving behaviour before they enter the influence of the pedestrian, and they tend to speed up when getting close to the opposite side of the sidewalk (to reach the goal and get their reward quicker). These issues should be addressed before the approach can be evaluated for use in real-world settings.

% Also, the robots have no explicit access to recent history (except for their last input to keep the input derivative low), nor do they use any predictions.

Yet, despite these limitations, the results show potential and encourage further exploration of this approach.

\section{Conclusion}
In this work, we provided a proof-of-concept for leveraging a human behaviour model in the planning and decision-making of a mobile robot. We trained a reinforcement learning algorithm to control a mobile robot in interactions with a Communication-Enabled Interaction model of a pedestrian. The results show that a basic RL agent successfully learned to interact with the CEI model. Furthermore, a risk-averse RL agent, that had access to the perceived risk of the CEI model during training, learned how to more effectively communicate its intention through its motion and thereby significantly lowered the risk perceived by the CEI model, and the effort the simulated pedestrian had to use to avoid the robot in a safe and acceptable manner.

\bibliographystyle{ieeetr}
\bibliography{references}

\begin{thebibliography}{10}

\bibitem{Siebinga2025}
O.~Siebinga, ``A model of the sidewalk salsa,'' in {\em IEEE International Conference on Systems, Man, and Cybernetics (SMC)}, (Vienna), IEEE, Oct. 2025.
\newblock arXiv:2412.04023 [cs].

\bibitem{Gehrke2023}
S.~R. Gehrke, C.~D. Phair, B.~J. Russo, and E.~J. Smaglik, ``{Observed sidewalk autonomous delivery robot interactions with pedestrians and bicyclists},'' {\em Transportation Research Interdisciplinary Perspectives}, vol.~18, no.~March, p.~100789, 2023.

\bibitem{Weinberg2023}
D.~Weinberg, H.~Dwyer, S.~E. Fox, and N.~Martelaro, ``{Sharing the Sidewalk: Observing Delivery Robot Interactions with Pedestrians during a Pilot in Pittsburgh, PA},'' {\em Multimodal Technologies and Interaction}, vol.~7, no.~5, 2023.

\bibitem{Dadvar2021}
M.~Dadvar, K.~Majd, E.~Oikonomou, G.~Fainekos, and S.~Srivastava, ``{Joint Communication and Motion Planning for Cobots},'' in {\em Proceedings - IEEE International Conference on Robotics and Automation}, pp.~4771--4777, sep 2022.

\bibitem{Chen2018}
Z.~Chen, C.~Jiang, and Y.~Guo, ``{Pedestrian-Robot Interaction Experiments in an Exit Corridor},'' {\em 2018 15th International Conference on Ubiquitous Robots, UR 2018}, pp.~29--34, 2018.

\bibitem{Fang_Mei_Yuan_Wang_Wang_Wang_2021}
B.~Fang, G.~Mei, X.~Yuan, L.~Wang, Z.~Wang, and J.~Wang, ``Visual slam for robot navigation in healthcare facility,'' {\em Pattern Recognition}, vol.~113, p.~107822, May 2021.

\bibitem{Möller_Furnari_Battiato_Härmä_Farinella_2021}
R.~Möller, A.~Furnari, S.~Battiato, A.~Härmä, and G.~M. Farinella, ``A survey on human-aware robot navigation,'' {\em Robotics and Autonomous Systems}, vol.~145, p.~103837, Nov. 2021.

\bibitem{Guest2021}
O.~Guest and A.~E. Martin, ``{How Computational Modeling Can Force Theory Building in Psychological Science},'' {\em Perspectives on Psychological Science}, vol.~16, no.~4, pp.~789--802, 2021.

\bibitem{Siebinga2023}
O.~Siebinga, A.~Zgonnikov, and D.~A. Abbink, ``{Modelling communication-enabled traffic interactions},'' {\em Royal Society Open Science}, vol.~10, may 2023.

\bibitem{Siebinga2024}
O.~Siebinga, A.~Zgonnikov, and D.~A. Abbink, ``A model of dyadic merging interactions explains human drivers’ behavior from control inputs to decisions,'' {\em PNAS Nexus}, vol.~3, p.~pgae420, 09 2024.

\bibitem{siebinga_modeling_2025}
O.~Siebinga, S.~H.~A. Mohammad, and A.~Zgonnikov, ``Modeling {Human} {Driver} {Behavior} {During} {Highway} {Merging} {Using} the {Communication}-{Enabled} {Interaction} {Framework},'' (Cluj), IEEE, June 2025.

\bibitem{Queiroz2024}
R.~Queiroz, D.~Sharma, R.~Caldas, K.~Czarnecki, S.~García, T.~Berger, and P.~Pelliccione, ``A driver-vehicle model for ads scenario-based testing,'' {\em IEEE Transactions on Intelligent Transportation Systems}, vol.~25, no.~8, pp.~8641--8654, 2024.

\bibitem{Li2018}
N.~Li, D.~W. Oyler, M.~Zhang, Y.~Yildiz, I.~Kolmanovsky, and A.~R. Girard, ``Game theoretic modeling of driver and vehicle interactions for verification and validation of autonomous vehicle control systems,'' {\em IEEE Transactions on Control Systems Technology}, vol.~26, no.~5, pp.~1782--1797, 2018.

\bibitem{Sadigh2018}
D.~Sadigh, N.~Landolfi, S.~S. Sastry, S.~A. Seshia, and A.~D. Dragan, ``{Planning for cars that coordinate with people: leveraging effects on human actions for planning and active information gathering over human internal state},'' {\em Autonomous Robots}, vol.~42, pp.~1405--1426, oct 2018.

\bibitem{Schwarting2019}
W.~Schwarting, A.~Pierson, J.~Alonso-Mora, S.~Karaman, and D.~Rus, ``{Social behavior for autonomous vehicles},'' {\em Proceedings of the National Academy of Sciences}, vol.~116, pp.~24972--24978, dec 2019.

\bibitem{helbing_social_1995}
D.~Helbing and P.~Molnár, ``Social force model for pedestrian dynamics,'' {\em Physical Review E}, vol.~51, pp.~4282--4286, May 1995.

\bibitem{Mombaur2008}
K.~Mombaur, J.~P. Laumond, and E.~Yoshida, ``{An optimal control model unifying holonomic and nonholonomic walking},'' {\em 2008 8th IEEE-RAS International Conference on Humanoid Robots, Humanoids 2008}, pp.~646--653, 2008.

\end{thebibliography}

\end{document}